\documentclass[runningheads]{llncs}
\usepackage[T1]{fontenc}
\usepackage{graphicx,verbatim}
\usepackage{xcolor}
\usepackage{booktabs}
\usepackage{tabularx}

\begin{document}
\title{Benchmarking Video Foundation Models for Remote Parkinson's Disease Screening}
\titlerunning{Benchmarking VFMs for PD Screening}
%

\author{Md Saiful Islam\inst{1} \and
Ekram Hossain\inst{1} \and
Abdelrahman Abdelkader\inst{1} \and
Tariq Adnan\inst{1} \and
Fazla Rabbi Mashrur\inst{1} \and
Sooyong Park\inst{1} \and
Praveen Kumar\inst{1} \and
Qasim Sudais\inst{1} \and
Natalia Chunga\inst{2} \and
Nami Shah\inst{3} \and
Jan Freyberg\inst{4} \and
Christopher Kanan\inst{1} \and
Ruth Schneider\inst{3} \and
Ehsan Hoque\inst{1}}
\authorrunning{M.S. Islam et al.}
%
\institute{University of Rochester, Rochester, NY, USA \and
Louisiana State University Health Sciences Center at Shreveport, USA \and
University of Rochester Medical Center, Rochester, NY, USA \and
University of Cambridge, Cambridge, UK\\
\email{mislam6@ur.rochester.edu}
}  
\maketitle              

\begin{abstract}
Video-based assessments offer a scalable pathway for remote Parkinson's disease (PD) screening. While traditional approaches rely on handcrafted features mimicking clinical scales, recent advances in video foundation models (VFMs) enable representation learning without task-specific customization. However, the comparative effectiveness of different VFM architectures across diverse clinical tasks remains poorly understood. We present a large-scale systematic study using a novel video dataset from 1,888 participants (727 with PD), comprising 32,847 videos across 16 standardized clinical tasks. We evaluate seven state-of-the-art VFMs -- including VideoPrism, V-JEPA, ViViT, and VideoMAE -- to determine their robustness in clinical screening. By evaluating frozen embeddings with a linear classification head, we demonstrate that task saliency is highly model-dependent: VideoPrism excels in capturing visual speech kinematics (no audio) and facial expressivity, while V-JEPA proves superior for upper-limb motor tasks. Notably, TimeSformer remains highly competitive for rhythmic tasks like finger tapping. Our experiments yield AUCs of 76.4–85.3\% and accuracies of 71.5–80.6\%. While high specificity (up to 90.3\%) suggests strong potential for ruling out healthy individuals, the lower sensitivity (43.2–57.3\%) highlights the need for task-aware calibration and integration of multiple tasks and modalities. Overall, this work establishes a rigorous baseline for VFM-based PD screening and provides a roadmap for selecting suitable tasks and architectures in remote neurological monitoring. Code and anonymized structured data are publicly available: \\
\textcolor{blue}{\scriptsize\url{https://anonymous.4open.science/r/parkinson\_video\_benchmarking-A2C5}}


\end{abstract}
\section{Introduction}
The sharply rising prevalence of Parkinson's disease (PD)~\cite{dorsey2018emerging} is creating an urgent need for scalable and accessible assessment tools. Traditionally, the diagnosis and severity assessment of PD rely on clinical evaluation and assessment of motor signs, including bradykinesia~\cite{bologna2023redefining}, rest tremor, and rigidity. The MDS-UPDRS~\cite{goetz2008movement}, which serves as the gold standard for assessment of motor signs, requires a trained specialist to rate the performance of standard motor tasks. However, geographic and financial barriers to specialized care prevent timely diagnosis in underserved populations~\cite{pearson2023care}.

\begin{figure}[t]
\centering
\includegraphics[width=\textwidth]{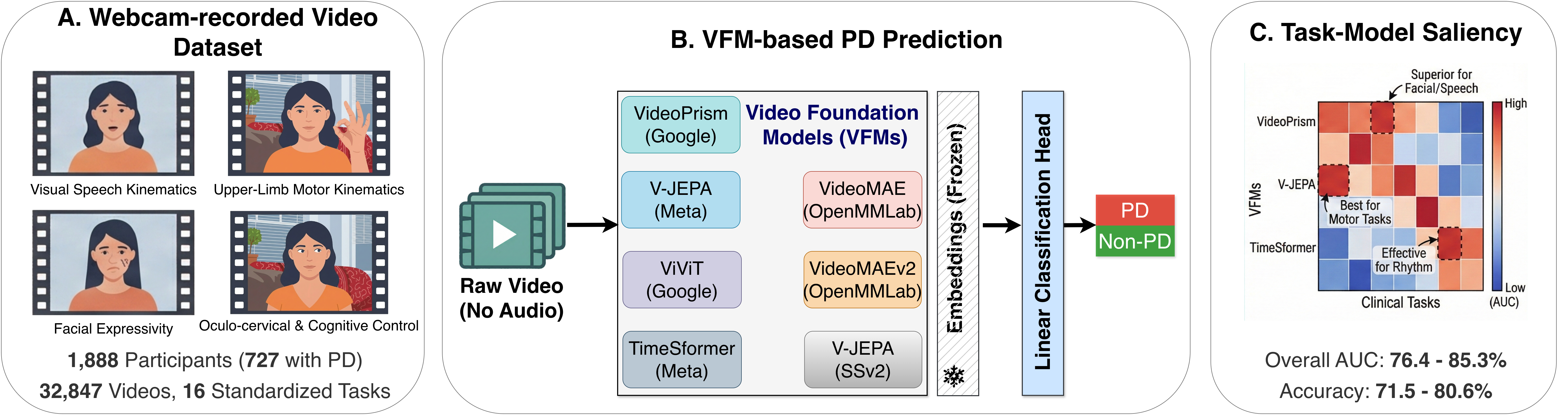}
\caption{\textbf{Overview of the benchmarking framework for PD screening using video foundation models (VFMs).} \textbf{(A) Video Dataset}: Our study utilizes a large-scale dataset from $1,888$ participants performing $16$ standardized clinical tasks. \textbf{(B) Evaluation Pipeline}: Raw video data is processed through a suite of state-of-the-art frozen VFMs to extract latent representations. These embeddings are evaluated using a task-specific linear classification head to differentiate between PD and non-PD participants. \textbf{(C) Task-Model Saliency}: Systematic evaluation to investigate architecture-specific strengths.} \label{fig:overview}
\end{figure}

Video-based assessments offer a scalable route by allowing subjects to perform standardized clinical tasks remotely using everyday devices (e.g., smartphones~\cite{bot2016mpower} and webcams~\cite{langevin2019park}) and obtain a preliminary PD risk screening. Computer-vision-based prior works focused on developing handcrafted features designed to mimic clinical observations (e.g., finger-tapping speed) and plugging them into machine learning models for PD screening~\cite{li2022detecting} and severity assessment~\cite{islam2023using}. In addition, the recent evolution of deep learning has introduced more sophisticated architectures, such as 3D ConvNets (e.g., I3D~\cite{carreira2017quo}), which can learn spatiotemporal representations directly from raw videos. However, these task-specific models either cannot generalize across other tasks without custom-engineered features or require large amounts of labeled data.

The emergence of Video Foundation Models (VFMs) may enable a paradigm shift in medical video analysis. VFMs are large-scale \emph{Vision Transformers}~\cite{dosovitskiy2021an} pre-trained on massive, heterogeneous datasets using self-supervised learning objectives. These models can learn versatile latent representations of visual dynamics, spanning fine-grained motion and high-level semantic interactions, without task-specific customization during pretraining. However, despite their broad capabilities in general action recognition~\cite{kay2017kinetics}, the comparative effectiveness of different VFM architectures across the specific clinical tasks used in PD screening remains poorly understood. These tasks vary in their requirements: some demand high-frequency temporal reasoning (e.g., finger tapping), while others require an understanding of subtle spatial deformations in facial expressivity or complex articulatory kinematics during speech. Our systematic benchmarking study addresses this gap by evaluating seven state-of-the-art VFMs -- VideoPrism~\cite{zhao2024videoprism}, V-JEPA2~\cite{assran2025v} (two variants), ViViT~\cite{arnab2021vivit}, VideoMAE~\cite{tong2022videomae}, VideoMAEv2~\cite{wang2023videomae}, and TimeSformer~\cite{bertasius2021space}.

By curating the largest webcam-recorded video dataset (32,847 videos from 1,888 participants; 727 with PD), we evaluate the ability of frozen embeddings from these models to differentiate between PD and non-PD across 16 standardized tasks (Fig.~\ref{fig:overview}). Our objective is to determine which architectural paradigms -- such as the semantic-visual distillation of VideoPrism, the latent-predictive world modeling of V-JEPA, or the divided space-time attention of TimeSformer -- are most salient for specific clinical tasks. By establishing a rigorous baseline for VFM-based PD screening, this work provides a roadmap for selecting suitable tasks and architectures for the next generation of remote video-based neurological monitoring systems.

\section{Methods}
\subsection{Dataset and Ethics}
The dataset was compiled from eight independent clinical and non-clinical studies (2017–2025), each approved by an \textbf{Institutional Review Board}. Participants were recruited via global and nationwide channels, including clinician referrals, social media, PD wellness centers, and research cohorts. We utilized a standardized web-based platform for recording tasks mimicking MDS-UPDRS assessments, performed either independently at home or under clinical supervision. PD status was determined through self-reporting (445) or clinical confirmation (282). Overall, the dataset comprises \textbf{32,847} total videos from \textbf{1,888} (991 female) individuals (see Fig.~\ref{fig:dataset} for age and clinical stage details). The cohort is predominantly White (1,366) with 68 Asian, 77 Black, and 27 individuals from other groups, while 350 participants did not disclose their race.

\subsection{Standardized Clinical Tasks and Domain Classification}
\begin{figure}[t]
\centering
\includegraphics[width=0.88\textwidth]{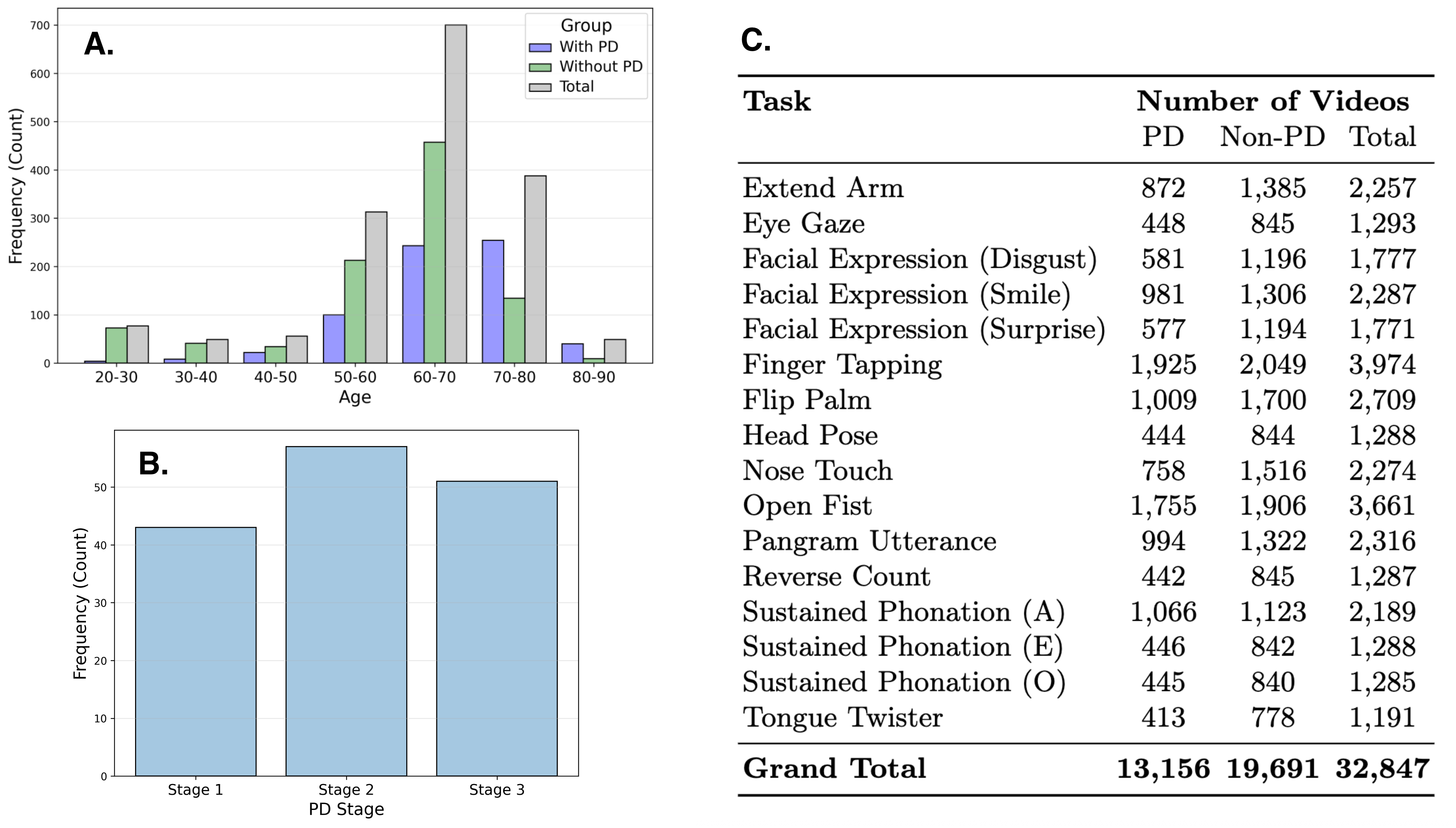}
\caption{\textbf{Dataset characteristics.} (A) The age distribution of participants categorized by PD status; (B) the distribution of clinical PD stage (Hoehn \& Yahr scale) when available (n = 158); and (C) the number of videos collected for each task.} \label{fig:dataset}
\end{figure}

The study evaluates performance across 16 standardized clinical tasks (see Fig.~\ref{fig:dataset}C.) inspired by the MDS-UPDRS~\cite{goetz2008movement} and other validated neurological assessments~\cite{langevin2019park}. We have organized these tasks into four broad clinical domains based on the physiological signs they are designed to elicit. To optimize quality and consistency in remote data collection setting, a video instruction demonstrating each specific task was provided to participants (available as \textbf{supplementary materials}).

\textbf{Upper-Limb Motor Kinematics:} This domain includes finger tapping, flip palm (hand pronation-supination), open fist (hand open-close), extending arms, and nose touching. These standard MDS-UPDRS items are utilized to assess bradykinesia~\cite{bologna2023redefining} (slowness of movement), hypokinesia~\cite{schilder2017terminology} (reduced amplitude), sequence effect (or decrement), and interruptions.

\textbf{Visual Speech Kinematics:} Tasks consist of uttering an English pangram (a sentence containing every letter of the alphabet), tongue twisters, and sustained phonation of vowels `a', `e', and `o'. While performed with audio, this study focuses on the \textbf{video-only} modality to capture articulatory movements of the jaw, lips, and tongue, which are frequently reduced or imprecise in PD.

\textbf{Facial Expressivity:} Tasks that require mimicking different facial expressions (smile, disgust, surprise) are used to evaluate hypomimia, characterized by a reduction in spontaneous and expressive facial movements~\cite{novotny2014automatic}.

\textbf{Oculomotor, Cervical, and Cognitive Control:} This domain assesses eye gaze (following an on-screen target), head pose (horizontal and vertical tilting), and reverse counting. These tasks evaluate a range of symptoms, including cervical dystonia~\cite{kilic2022current}, oculomotor control, and cognitive-motor interference -- a task requiring mental flexibility that is often impaired in PD~\cite{leone2017cognitive}.





\subsection{Video Foundation Model Architectures}

We evaluate seven state-of-the-art VFMs spanning diverse architectures and pretraining objectives. Together, they provide a broad benchmark across temporal modeling (e.g., divided attention vs. factorized encoding) and supervision strategies (e.g., pixel reconstruction, latent prediction, and language alignment).

\begin{itemize}
    \item \textbf{VideoPrism~\cite{zhao2024videoprism}:} Employs a two-stage pre-training strategy involving video-text contrastive learning on 36M pairs followed by masked video modeling with global-local distillation. This hybrid approach allows the model to capture both high-level semantic language cues and fine-grained visual motion.

    \item \textbf{V-JEPA2~\cite{assran2025v} and V-JEPA2 (SSv2):} Utilizes a \textbf{J}oint \textbf{E}mbedding \textbf{P}redictive \textbf{A}rchitecture that forecasts the latent representations of masked regions from unmasked context rather than reconstructing raw pixels. We include a version fine-tuned on the Something-Something v2 (SSv2~\cite{goyal2017something}) dataset to assess if motion-centric supervision enhances clinical motor task performance.

    \item \textbf{TimeSformer~\cite{bertasius2021space} and ViViT~\cite{arnab2021vivit}:} These models utilize factorized attention mechanisms. TimeSformer uses ``divided space-time attention'' to sequentially process spatial and temporal features, while ViViT factorizes the spatial and temporal encoders to handle complexity of video transformers.

    \item \textbf{VideoMAE~\cite{tong2022videomae} and VideoMAEv2~\cite{wang2023videomae}:} These models use masked autoencoding to reconstruct the original pixels of highly masked (up to 90\%) video frames. VideoMAEv2 further scales this approach using dual masking to efficiently learn detailed visual structures across massive datasets.
\end{itemize}

\subsection{Experimental Protocol}

To evaluate foundation models for PD risk screening, we use a ``frozen-backbone'' protocol: VFM weights remain fixed after pretraining and are used only as feature extractors. We resize input videos and sample frames at 15 FPS, following model-specific requirements, to obtain spatiotemporal embeddings. \textbf{Pretrained model weights and implementation are available in our public repository.}

A neural network featuring one hidden layer (ReLU activation, random dropout) and a linear classification head is trained on these embeddings to minimize binary cross-entropy loss against ground-truth PD/non-PD labels. The hidden layer dimension is treated as a tunable hyperparameter. Data is randomly partitioned into training ($60\%$), validation ($20\%$), and test ($20\%$) folds based on unique participants to ensure \textbf{evaluation on unseen subjects}. We optimize hyperparameters, including learning rate, hidden nodes, and dropout, using Weights \& Biases, allocating equal compute time across all tasks and VFMs (script for hyperparameter search is provided in our repository). The final model for each task is selected based on peak validation set area under the receiver operating characteristic curve (AUC). When feasible, results are reported as the mean and $95\%$ confidence interval (CI) obtained across $30$ different random seeds. 

Additionally, we explore multi-view training, where a secondary linear head aggregates likelihoods from individual views, and assess oversampling strategies (e.g., RandomOverSampler~\cite{lemaavztre2017imbalanced}) to address class imbalance. All experiments are conducted locally on a 32-core AMD Ryzen Threadripper PRO 5975WX workstation equipped with dual NVIDIA RTX A6000 (each with 48 GB vRAM) and 256 GB RAM. Local deployment guarantees that \textbf{sensitive patient videos are processed on-premises} without exposure to cloud or third-party servers.
\section{Results and Discussions}

\begin{table*}[t]
\centering
\caption{\textbf{Performance of Best Models across standardized tasks.} Results are reported for the model best performing on the validation set (AUC) on each task. Bold denotes the overall best result, and $\pm$ indicates $95\%$ confidence intervals.}
\label{tab:single_view_benchmark}
\resizebox{0.95\columnwidth}{!}{%
\begin{tabular}{l|l|c|c|c|c|c|c}
\toprule
\textbf{Task} & \textbf{Best Model} & \textbf{AUC} & \textbf{Accuracy} & \textbf{Sensitivity} & \textbf{Specificity} & \textbf{PPV} & \textbf{NPV} \\
\midrule
Extend Arm & V-JEPA2-SSv2 & $83.2 \pm 0.1$ & $78.2 \pm 0.3$ & $55.8 \pm 1.4$ & $87.8 \pm 0.6$ & $66.4 \pm 0.8$ & $82.3 \pm 0.4$ \\
Eye Gaze & VideoPrism & $81.0 \pm 0.1$ & $75.2 \pm 0.3$ & $\mathbf{57.3 \pm 1.1}$ & $83.9 \pm 0.8$ & $63.6 \pm 0.9$ & $80.1 \pm 0.3$ \\
Facial Expression Disgust & V-JEPA2 & $79.2 \pm 0.1$ & $78.7 \pm 0.3$ & $\mathbf{57.3 \pm 0.8}$ & $87.8 \pm 0.4$ & $66.7 \pm 0.6$ & $82.9 \pm 0.2$ \\
Facial Expression Smile & VideoPrism & $80.3 \pm 0.3$ & $74.3 \pm 0.3$ & $51.4 \pm 2.2$ & $84.1 \pm 1.1$ & $58.3 \pm 0.9$ & $80.3 \pm 0.5$ \\
Facial Expression Surprise & V-JEPA2-SSv2 & $80.2 \pm 0.1$ & $77.4 \pm 0.4$ & $55.3 \pm 1.4$ & $86.6 \pm 0.8$ & $63.5 \pm 1.0$ & $82.3 \pm 0.4$ \\
Finger Tapping & TimeSformer & $76.4 \pm 0.3$ & $73.0 \pm 0.4$ & $43.2 \pm 1.2$ & $85.8 \pm 0.9$ & $57.0 \pm 1.1$ & $77.8 \pm 0.2$ \\
Flip Palm & V-JEPA2-SSv2 & $\mathbf{85.3 \pm 0.2}$ & $\mathbf{80.6 \pm 0.3}$ & $56.2 \pm 0.9$ & $\mathbf{90.3 \pm 0.5}$ & $\mathbf{69.7 \pm 0.9}$ & $\mathbf{83.9 \pm 0.2}$ \\
Head Pose & VideoPrism & $79.3 \pm 0.1$ & $71.7 \pm 0.4$ & $52.5 \pm 1.1$ & $81.0 \pm 1.0$ & $57.3 \pm 0.9$ & $77.9 \pm 0.2$ \\
Nose Touch & V-JEPA2-SSv2 & $83.0 \pm 0.1$ & $79.3 \pm 0.2$ & $54.2 \pm 1.0$ & $89.3 \pm 0.3$ & $66.8 \pm 0.5$ & $83.1 \pm 0.3$ \\
Open Fist & VideoPrism & $84.3 \pm 0.1$ & $76.8 \pm 0.3$ & $54.6 \pm 1.9$ & $85.7 \pm 0.5$ & $60.5 \pm 0.4$ & $82.5 \pm 0.5$ \\
Pangram Utterance & VideoPrism & $81.3 \pm 0.2$ & $75.5 \pm 0.4$ & $50.1 \pm 1.0$ & $86.8 \pm 0.9$ & $63.2 \pm 1.4$ & $79.6 \pm 0.2$ \\
Reverse Count & VideoPrism & $80.0 \pm 0.1$ & $73.2 \pm 0.3$ & $52.8 \pm 0.9$ & $83.0 \pm 0.5$ & $60.0 \pm 0.6$ & $78.5 \pm 0.3$ \\
Sustained Phonation A & VideoPrism & $79.2 \pm 0.2$ & $71.5 \pm 0.5$ & $54.8 \pm 2.9$ & $79.7 \pm 1.5$ & $57.2 \pm 1.2$ & $78.5 \pm 0.9$ \\
Sustained Phonation E & VideoPrism & $80.7 \pm 0.1$ & $73.1 \pm 0.3$ & $51.9 \pm 1.0$ & $83.4 \pm 0.6$ & $60.5 \pm 0.6$ & $78.0 \pm 0.3$ \\
Sustained Phonation O & VideoPrism & $81.0 \pm 0.1$ & $73.4 \pm 0.3$ & $56.3 \pm 1.2$ & $81.7 \pm 0.5$ & $60.1 \pm 0.5$ & $79.3 \pm 0.4$ \\
Tongue Twister & VideoPrism & $77.1 \pm 0.2$ & $73.4 \pm 0.3$ & $54.0 \pm 1.2$ & $82.4 \pm 0.7$ & $58.9 \pm 0.6$ & $79.4 \pm 0.4$ \\
\bottomrule
\end{tabular}%
}
\end{table*}

\subsection{Task-wise Saliency Patterns and Clinical Implications}
Performance results across the 16 tasks (see Table~\ref{tab:single_view_benchmark}) indicate that upper-limb motor maneuvers exhibit the highest saliency. Specifically, \texttt{Flip Palm} and \texttt{Open Fist} yielded the highest AUCs (95\% CI: $85.3 \pm 0.2\%$ and $84.3 \pm 0.1\%$), with \texttt{Flip Palm} achieving a peak specificity of $90.3 \pm 0.5\%$. These findings align with clinical literature identifying limb bradykinesia as assessed with rapid alternating movements as a high-yield diagnostic markers during clinical assessments~\cite{sarasso2024neural}.

Conversely, fine-motor tasks like \texttt{Finger Tapping} and vocal tasks like \texttt{Tongue Twister} showed lower saliency (AUCs $<78$). This likely reflects the difficulty of capturing subtle movements or dysarthric shifts~\cite{di2022voice} via general video-based embeddings (we did not use the audio). While models showed relatively higher sensitivity in facial tasks (i.e., \texttt{Eye Gaze} and \texttt{Disgust}), overall performance suggests that self-supervised models require enhanced temporal resolution or localized attention to match expert clinical observation. In addition, domain-specific feature extraction~\cite{adnan2025ai} or advanced architecture design may be necessary to outperform the VFM baselines established here.
Finally, the discrepancy between high AUC and low sensitivity suggests non-optimal decision thresholds. Clinical utility in screening will require improved model calibration (i.e., post-hoc scaling of predicted probabilities~\cite{platt1999probabilistic}). Prior literature also suggests that integrating multiple tasks and modalities can significantly improve performance~\cite{islam2025accessible}.

\subsection{Architecture-Specific Strengths Across Clinical Domains}
\begin{figure}[t]
\centering
\includegraphics[width=\textwidth]{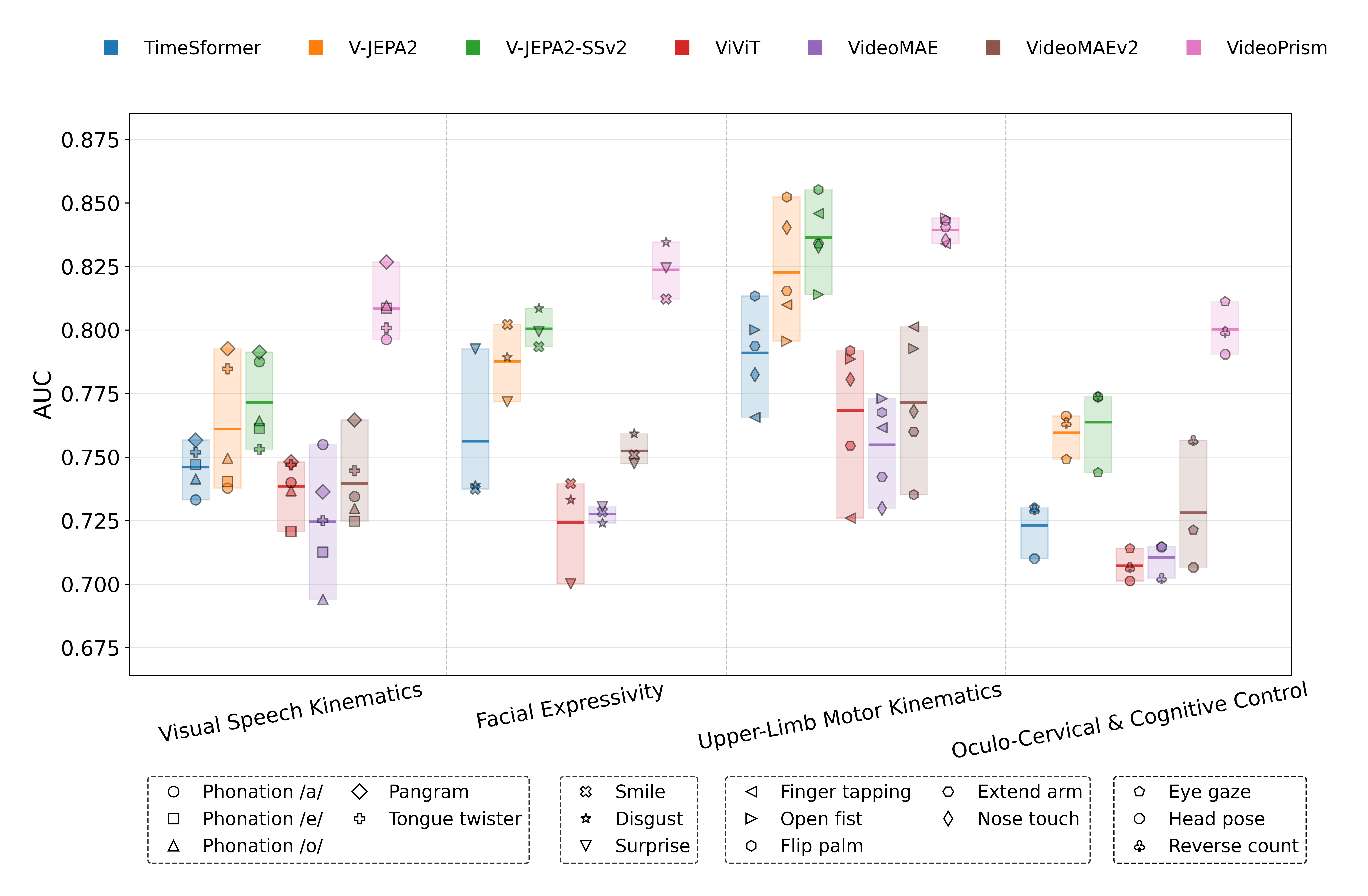}
\caption{\textbf{Comparative performance of VFMs across broad clinical domains.} This plot organizes results by broad clinical category along the x-axis, with color-coded boxes showing the performance distribution of each VFM within that domain (horizontal lines showing mean). Individual data points, distinguished by unique marker shapes (defined in the bottom legend), represent the specific AUC achieved by a model on a distinct clinical task. High-level analysis reveals that the ``Upper-Limb Motor Kinematics'' domain generally yields the highest predictive performance. At the task level, distinct model specializations emerge: \colorbox{pink}{VideoPrism} shows superior consistency across ``Facial Expressivity,'' ``Visual Speech Kinematics,'' and ``Oculo-Cervical \& Cognitive Control''. In contrast, V-JEPA models dominate in ``Upper-Limb Motor Kinematics,'' achieving the highest overall scores on tasks like flip-palm. \colorbox{green}{V-JEPA-SSv2} generally outperforms \colorbox{orange}{V-JEPA}, suggesting additional benefits of training on the SSv2 dataset.} \label{fig1}
\end{figure}

The comparative evaluation of VFM architectures across diverse domains highlights specific structural advantages for PD screening. \textbf{VideoPrism} demonstrated strong generalization, ranking first across 10 of 16 tasks, while providing competitive performance in other tasks. Its superiority is most pronounced in \texttt{Visual Speech Kinematics} and \texttt{Facial Expressivity} domains, where the model needed to capture the subtle, low-amplitude spatio-temporal features associated with hypomimia and dysarthria. In addition, given its robust cross-domain performance, we recommend VideoPrism as the primary baseline for general-purpose PD screening or research focused on orofacial manifestations~\cite{friedlander2009parkinson}.

For \texttt{Upper-Limb Motor Kinematics}, specifically tasks involving high-amplitude movements like \texttt{Flip Palm}, the \textbf{V-JEPA2} family proved most effective. Notably, V-JEPA2-SSv2 consistently outperformed the base V-JEPA2 model in tasks requiring precise limb coordination and trajectory tracking, such as \texttt{Extend Arm} and \texttt{Nose Touch}. This performance gap suggests that the additional fine-tuning on the Something-Something v2 (SSv2) dataset -- which is heavily oriented toward human-object interactions and directional motion -- likely enhanced the model's ability to represent complex temporal dynamics and motion causality. Consequently, V-JEPA2-SSv2 may serve as the optimal baseline for research targeting bradykinesia and motor coordination.

Ultimately, foundation model selection should depend on the clinical domain. Although VideoPrism provides the strongest baseline, VJEPA2-SSv2's motion-focused pretraining better captures rhythmic irregularities in gross motor symptoms. For fine-motor tasks such as \texttt{Finger Tapping}, where \textbf{TimeSformer} performed best, current VFMs may benefit from architectures that emphasize high-resolution temporal attention rather than broad semantic embeddings.

\subsection{Multi-View and Oversampling Ablations}


We evaluated multi-view aggregation and oversampling to handle videos longer than a single view supports and to mitigate class imbalance. The single-view configuration without oversampling generally provided robust results, achieving a mean accuracy of $74.97\%$ (std: $3.12\%$) and a mean AUC of $80.54\%$ (std: $2.42\%$) across all the 16 tasks. While multi-view training was expected to improve performance by capturing more temporal context, it did not significantly outperform single-view metrics (p>0.70 for both accuracy and AUC), suggesting that salient diagnostic features maybe often captured within short segments. An alternative explanation is we need more complex architecture to integrate multiple views, as they substantially increase feature dimension (which remains out of scope for this study). Furthermore, oversampling strategies were found to slightly (non-significant) degrade mean accuracy ($74.51\%$; std = $2.99\%$) and AUC ($79.83\%$; std = $2.82\%$), indicating that the frozen VFM embeddings are already robust enough to handle existing class imbalances, and artificial data expansion may introduce noise into the latent feature space.

\subsection{Limitations}
While this work establishes a rigorous baseline, several limitations remain. The ``frozen evaluation'' protocol, although ideal for benchmarking the inherent representational power of VFMs, does not explore the performance ceilings achievable through task-specific fine-tuning or parameter-efficient methods such as LoRA~\cite{hu2022lora}. Moreover, to safeguard patient privacy and avoid sending sensitive videos to third-party services, we restricted experiments to open-weight models that can be deployed locally. As the VFM ecosystem evolves, future work should evaluate newer architectures as they become available. In addition, while self-reported PD diagnosis labels helped scale up the study, they may introduce some label noise. Finally, despite a diverse, open recruitment process, the cohort remains predominantly white, which may limit the generalizability of our findings to more diverse populations.
\section{Conclusion}
This study provides a rigorous benchmark of video foundation models for remote Parkinson’s disease screening. Our results suggest that frozen embeddings from large pretrained models can capture clinically meaningful signs, even without task-specific fine-tuning. We evaluated seven state-of-the-art VFMs on 16 standardized tasks. The results show that model choice should match the physiological domain. Motion-predictive models such as V-JEPA work best for upper-limb motor kinematics. In contrast, semantic-visual models such as VideoPrism are better for subtle facial and speech-related dynamics. Overall, these findings offer a clear roadmap for video-based movement disorder assessment. Future remote monitoring tools may benefit from modular VFM-based pipelines, alongside carefully designed handcrafted features.

\bibliographystyle{splncs04}
\bibliography{main}
\end{document}